\documentclass[journal,11pt]{IEEEtran}
\usepackage{soul}
\usepackage{color}
\usepackage{amsmath}
\usepackage{amssymb}
\usepackage{graphicx}
\usepackage[scaled=0.88]{helvet}
 
\usepackage[T1]{fontenc}

\begin{document}

\title{Generative Adversarial Networks: An Overview}

\author{Antonia~Creswell\IEEEauthorrefmark{4}, 
Tom~White\IEEEauthorrefmark{5}, Vincent~Dumoulin\IEEEauthorrefmark{3},~
Kai~Arulkumaran\IEEEauthorrefmark{4},~Biswa~Sengupta\IEEEauthorrefmark{2}\IEEEauthorrefmark{4}~
and~Anil~A~Bharath\IEEEauthorrefmark{4},~\IEEEmembership{Member~IEEE} \\ 
\IEEEauthorrefmark{4} BICV Group, Dept. of Bioengineering, Imperial College London\\
\IEEEauthorrefmark{5} School of Design, Victoria University of Wellington, New Zealand\\
\IEEEauthorrefmark{3} MILA, University of Montreal, Montreal H3T 1N8\\
\IEEEauthorrefmark{2} Cortexica Vision Systems Ltd., London, United Kingdom\\}

%
%

\markboth{Submitted to IEEE-SPM, April~2017}%
{Shell \MakeLowercase{\textit{et al.}}: Generative Adversarial Networks: An Overview}
%



\maketitle

\begin{abstract}
Generative adversarial networks (GANs) provide a way to learn deep representations without extensively annotated training data. They achieve this through deriving backpropagation signals through a competitive process involving a pair of networks. The representations that can be learned by GANs may be used in a variety of applications, including image synthesis, semantic image editing, style transfer, image super-resolution and classification. The aim of this review paper is to provide an overview of GANs for the signal processing community, drawing on familiar analogies and concepts where possible. In addition to identifying different methods for training and constructing GANs, we also point to remaining challenges in their theory and application.
\end{abstract}


%
\IEEEpeerreviewmaketitle

\section{Introduction}
\begin{figure*}[h!]
\centering
\includegraphics[width=12cm]{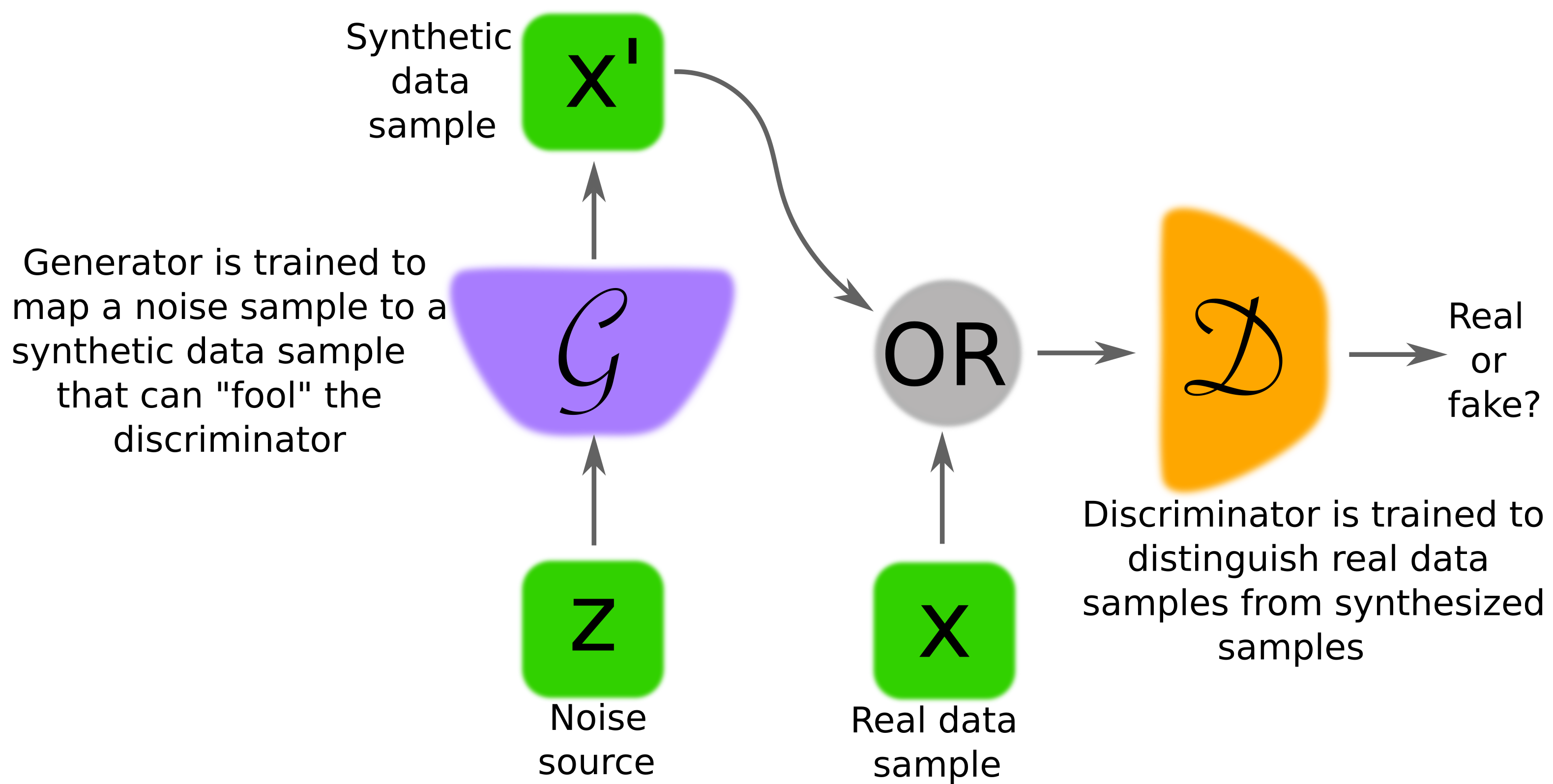}
\caption{In this figure, the two models which are learned during the training process for a GAN are the discriminator ($\mathcal{D}$) and the generator ($\mathcal{G}$). These are typically implemented with neural networks, but they could be implemented by any form of differentiable system that maps data from one space to another; see text for details.}
\label{fig:plaingan}
\end{figure*}

\IEEEPARstart{G}{enerative} adversarial networks (GANs) are an emerging technique for both semi-supervised and unsupervised learning. They achieve this through implicitly modelling high-dimensional distributions of data. Proposed in 2014 \cite{goodfellow2014generative}, they can be characterized by training a pair of networks in competition with each other. A common analogy, apt for visual data, is to think of one network as an art forger, and the other as an art expert. The forger, known in the GAN literature as the generator, $\mathcal{G}$, creates forgeries, with the aim of making realistic images. The expert, known as the discriminator, $\mathcal{D}$, receives both forgeries and real (authentic) images, and aims to tell them apart (see Fig.~\ref{fig:plaingan}). Both are trained simultaneously, and in competition with each other.

Crucially, the generator has no direct access to real images - the only way it learns is through its interaction with the discriminator. The discriminator has access to both the synthetic samples and samples drawn from the stack of real images. The error signal to the discriminator is provided through the simple ground truth of knowing whether the image came from the real stack or from the generator. The same error signal, via the discriminator, can be used to train the generator, leading it towards being able to produce forgeries of better quality. 

The networks that represent the generator and discriminator are typically implemented by multi-layer networks consisting of convolutional and/or fully-connected layers. The generator and discriminator networks must be differentiable, though it is not necessary for them to be directly invertible. If one considers the generator network as mapping from some representation space, called a latent space, to the space of the data (we shall focus on images), then we may express this more formally as $\mathcal{G}:\mathcal{G}(\mathbf{z})\rightarrow R^{|\mathbf{x}|}$, where $\mathbf{z} \in R^{|\mathbf{z}|}$ is a sample from the latent space, $\mathbf{x} \in R^{|\mathbf{x}|}$ is an image and $| \cdot |$ denotes the number of dimensions.

In a basic GAN, the discriminator network, $\mathcal{D}$, may be similarly characterized as a function that maps from image data to a probability that the image is from the real data distribution, rather than the generator distribution: $\mathcal{D}:\mathcal{D}(\mathbf{x})\rightarrow (0,1)$. For a fixed generator, $\mathcal{G}$, the discriminator, $\mathcal{D}$, may be trained to classify images as either being from the training data (real, close to 1) or from a fixed generator (fake, close to 0). When the discriminator is optimal, it may be frozen, and the generator, $\mathcal{G}$, may continue to be trained so as to lower the accuracy of the discriminator. If the generator distribution is able to match the real data distribution perfectly then the discriminator will be maximally confused, predicting $0.5$ for all inputs. In practice, the discriminator might not be trained until it is optimal; we explore the training process in more depth in Section \ref{sec:trainingGAN}.



On top of the interesting academic problems related to training and constructing GANs, the 
motivations behind training GANs may not necessarily be the generator or the discriminator {\em per se}: 
the representations embodied by either of the pair of networks can be used in a variety of subsequent tasks. We explore the applications of these representations in Section~\ref{sec:Apps}.

\section{Preliminaries}
\subsection{Terminology}
Generative models learn to capture the statistical distribution of training data, allowing us to synthesize samples from the learned distribution. On top of synthesizing novel data samples, which may be used for downstream tasks such as semantic image editing \cite{zhu2016generative}, data augmentation \cite{bousmalis2016unsupervised} and style transfer \cite{zhu2017unpaired}, we are also interested in using the representations that such models learn for tasks such as classification \cite{radford2015unsupervised} and image retrieval \cite{creswell2016adversarial}.


We occasionally refer to fully connected and convolutional layers of deep networks; these are generalizations of perceptrons or of spatial filter banks with non-linear post-processing. In all cases, the network weights are learned through backpropagation \cite{lecun2015deep}.

\subsection{Notation}
The GAN literature generally deals with multi-dimensional vectors, and often represents vectors in a probability space by italics (e.g. latent space is $z$). In the field of signal processing, it is common to represent vectors by bold lowercase symbols, and we adopt this convention in order to emphasize the multi-dimensional nature of variables. Accordingly, we will commonly refer to $p_{data}(\mathbf{x})$ as representing the probability density function over a random vector $\mathbf{x}$ which lies in $R^{|\mathbf{x}|}$.  We will use $p_g(\mathbf{x})$ to denote the distribution of the vectors produced by the generator network of the GAN. We use the calligraphic symbols $\mathcal{G}$ and $\mathcal{D}$ to denote the generator and discriminator networks, respectively. Both networks have sets of parameters (weights), ${\Theta}_D$ and ${\Theta}_G$, that are learned through optimization, during training.

As with all deep learning systems, training requires that we have some clear objective function. Following the usual notation, we use $J_{G}(\Theta_G;\Theta_D)$ and $J_{D}(\Theta_D;\Theta_G)$ to refer to the objective functions of the generator and discriminator, respectively. The choice of notation reminds us that the two objective functions are in a sense co-dependent on the evolving parameter sets $\Theta_G$ and $\Theta_D$ of the networks as they are iteratively updated. We shall explore this further in Section \ref{sec:trainingGAN}. Finally, note that multidimensional gradients are used in the updates; we use $\nabla_{\Theta_G}$ to denote the gradient operator with respect to the weights of the generator parameters, and $\nabla_{\Theta_D}$ to denote the gradient operator with respect to the weights of the discriminator. The expected gradients are indicated by the notation $\mathbb{E}\nabla_{\bullet}$.

\subsection{Capturing Data Distributions}
A central problem of signal processing and statistics is that {\em of density estimation:} obtaining a representation -- implicit or explicit, parametric or non-parametric -- of data in the real world. This is the key motivation behind GANs.  In the GAN literature, the term {\em data generating distribution} is often used to refer to the underlying probability density or probability mass function of observation data. 
GANs learn through implicitly computing some sort of similarity between the distribution of a candidate model and the distribution corresponding to real data.  

Why bother with density estimation at all? The answer lies at the heart of -- arguably -- many problems of visual inference, including image categorization, visual object detection and recognition, object tracking and object registration. In principle, through Bayes' Theorem, all inference problems of computer vision can be addressed through estimating conditional density functions, possibly indirectly in the form of a model which learns the joint distribution of variables of interest and the observed data. The difficulty we face is that likelihood functions for high-dimensional, real-world image data are difficult to construct. Whilst GANs don't explicitly provide a way of evaluating density functions, for a generator-discriminator pair of suitable capacity, the generator implicitly captures the distribution of the data.  

\begin{figure*}[h!]
\centering
\includegraphics[width=12cm]{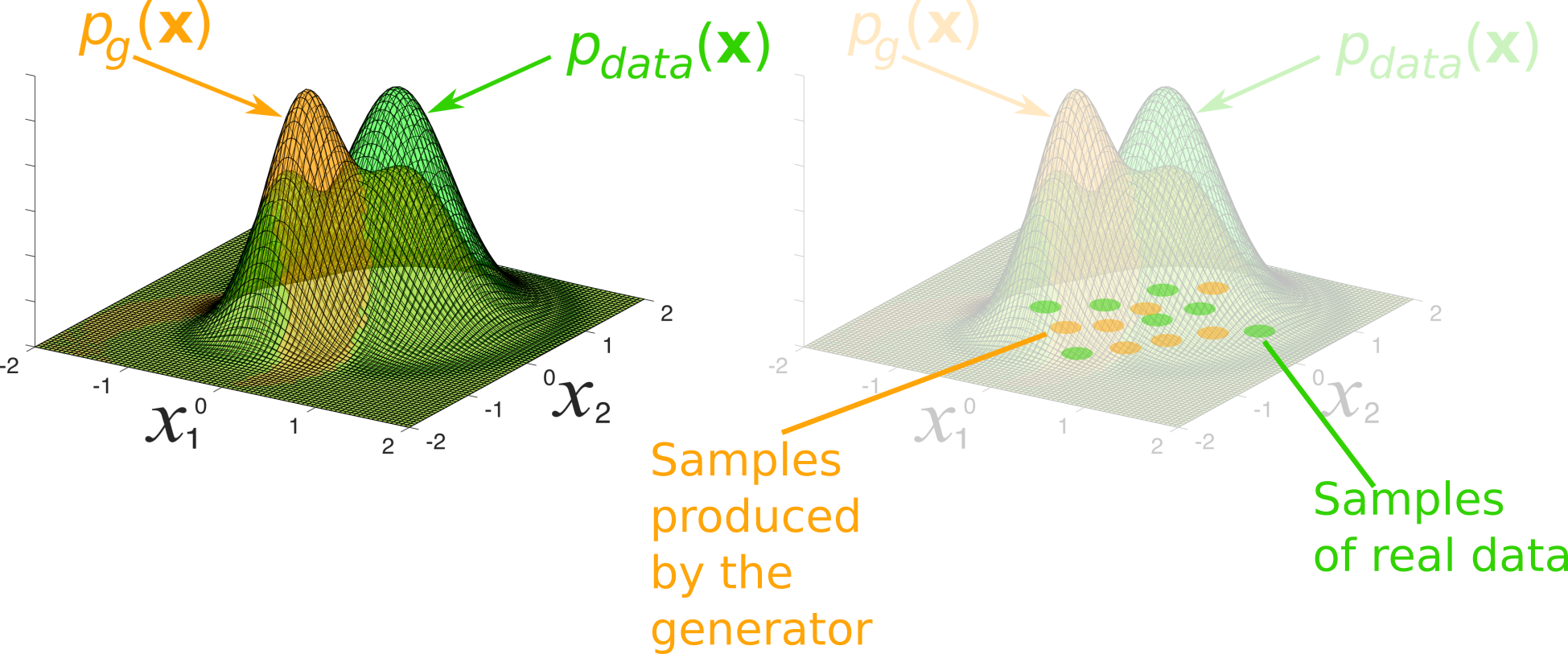}
\caption{During GAN training, the generator is encouraged to produce a distribution of samples, $p_g(\mathbf{x})$ to match that of real data, $p_{data}(\mathbf{x})$. For an appropriately parametrized and trained GAN, these distributions will be nearly identical. The representations embodied by GANs are captured in the learned parameters (weights) of the generator and discriminator networks.}
\label{fig:pgvspdata}
\end{figure*}


\subsection{Related Work}
\label{sec:PreliminariesRelWork}
One may view the principles of generative models by making comparisons with standard techniques in signal processing and data analysis. For example, signal processing makes wide use of the idea of representing a signal as the weighted combination of basis functions. Fixed basis functions underlie standard techniques such as Fourier-based and wavelet representations. Data-driven approaches to constructing basis functions can be traced back to the Hotelling \cite{hotelling1933analysis} transform, rooted in Pearson's observation that principal components minimize a reconstruction error according to a minimum squared error criterion.  Despite its wide use, standard Principal Components Analysis (PCA) does not have an overt statistical model for the observed data, though it has been shown that the bases of PCA may be derived as a maximum likelihood parameter estimation problem. 

Despite wide adoption, PCA itself is limited -- the basis functions emerge as the eigenvectors of the covariance matrix over observations of the input data, and the mapping from the representation space back to signal or image space is linear. So, we have both a shallow and a linear mapping, limiting the complexity of the model, and hence of the data, that can be represented. 
 
Independent Components Analysis (ICA) provides another level up in sophistication, in which the signal components no longer need to be orthogonal; the mixing coefficients used to blend components together to construct examples of data are merely considered to be statistically independent. ICA has various formulations that differ in their objective functions used during estimating signal components, or in the generative model that expresses how signals or images are generated from those components. A recent innovation explored through ICA is noise contrastive estimation (NCE); this may be seen as approaching the spirit of GANs \cite{goodfellow2014distinguishability}: the objective function for learning independent components compares a statistic applied to noise with that produced by a candidate generative model \cite{gutmann2010noise}. The original NCE approach did not include updates to the generator.


What other comparisons can be made between GANs and the standard tools of signal processing? For PCA, ICA, Fourier and wavelet representations, the latent space of GANs is, by analogy, the coefficient space of what we commonly refer to as transform space. What sets GANs apart from these standard tools of signal processing is the level of complexity of the models that map vectors from latent space to image space. Because the generator networks contain non-linearities, and can be of almost arbitrary depth, this mapping -- as with many other deep learning approaches -- can be extraordinarily complex.
 
With regard to deep image-based models, modern approaches to generative image modelling can be grouped into explicit density models and implicit density models. Explicit density models are either tractable (change of variables models, autoregressive models) or intractable (directed models trained with variational inference, undirected models trained using Markov chains). Implicit density models capture the statistical distribution of the data through a generative process which makes use of either ancestral sampling \cite{bengio2013generalized} or Markov chain-based sampling. GANs fall into the directed implicit model category. A more detailed overview and relevant papers can be found in Ian Goodfellow's NIPS 2016 tutorial \cite{goodfellow2016nips}.





\section{GAN Architectures}
\label{sec:GANArch}
\subsection{Fully Connected GANs}
The first GAN architectures used fully connected neural networks for both the generator and discriminator \cite{goodfellow2014generative}. This type of architecture was applied to relatively simple image datasets, namely MNIST (hand written digits), CIFAR-10 (natural images) and the Toronto Face Dataset (TFD).


\subsection{Convolutional GANs}
\label{sec:ConvGANArch}

Going from fully-connected to convolutional neural networks is a natural extension, given that CNNs are extremely well suited to image data. Early experiments conducted on CIFAR-10 suggested that it was more difficult to train generator and discriminator networks using CNNs with the same level of capacity and representational power as the ones used for supervised learning.

The Laplacian pyramid of adversarial networks (LAPGAN) \cite{denton2015deep} offered one solution to this problem, by decomposing the generation process using multiple scales: a ground truth image is itself decomposed into a Laplacian pyramid, and a conditional, convolutional GAN is trained to produce each layer given the one above.

Additionally, Radford et al. \cite{radford2015unsupervised} proposed a family of network architectures called DCGAN (for ``deep convolutional GAN'') which allows training a pair of deep convolutional generator and discriminator networks. DCGANs make use of strided and fractionally-strided convolutions which allow the spatial down-sampling and up-sampling operators to be learned during training. These operators handle the change in sampling rates and locations, a key requirement in mapping from image space to possibly lower-dimensional latent space, and from image space to a discriminator. Further details of the DCGAN architecture and training are presented in Section~\ref{sec:trainingTricks}.

As an extension to synthesizing images in 2D, Wu et al. \cite{wu2016learning} presented GANs that were able to synthesize 3D data samples using \textit{volumetric} convolutions. Wu et al. \cite{wu2016learning} synthesized novel objects including chairs, table and cars; in addition, they also presented a method to map from 2D image images to 3D versions of objects portrayed in those images.

\subsection{Conditional GANs}
\label{sec:CondGANArch}
\begin{figure*}[h!]
\centering
\includegraphics[width=8cm]{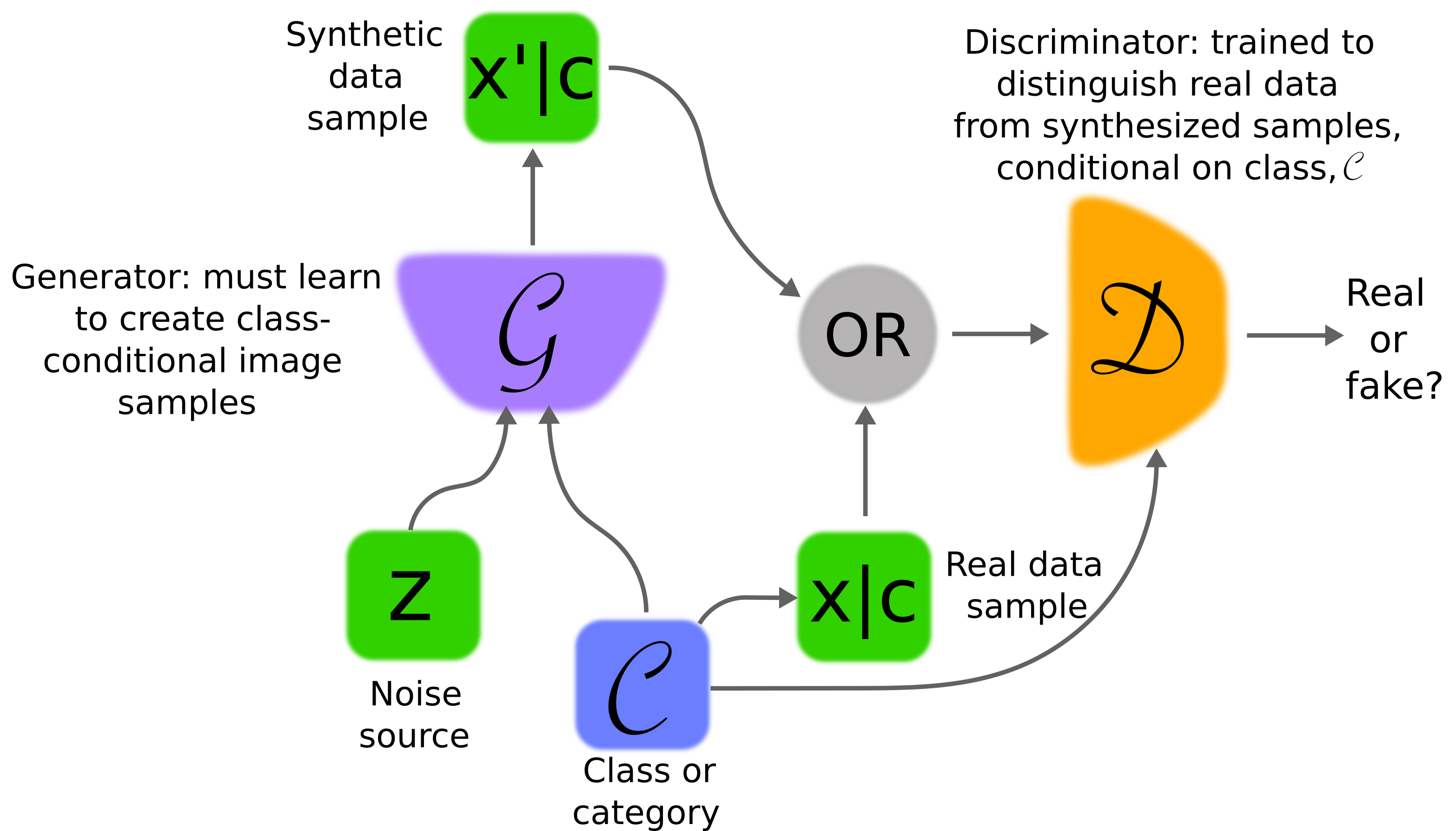}
\hspace{0.2cm}
\includegraphics[width=8cm]{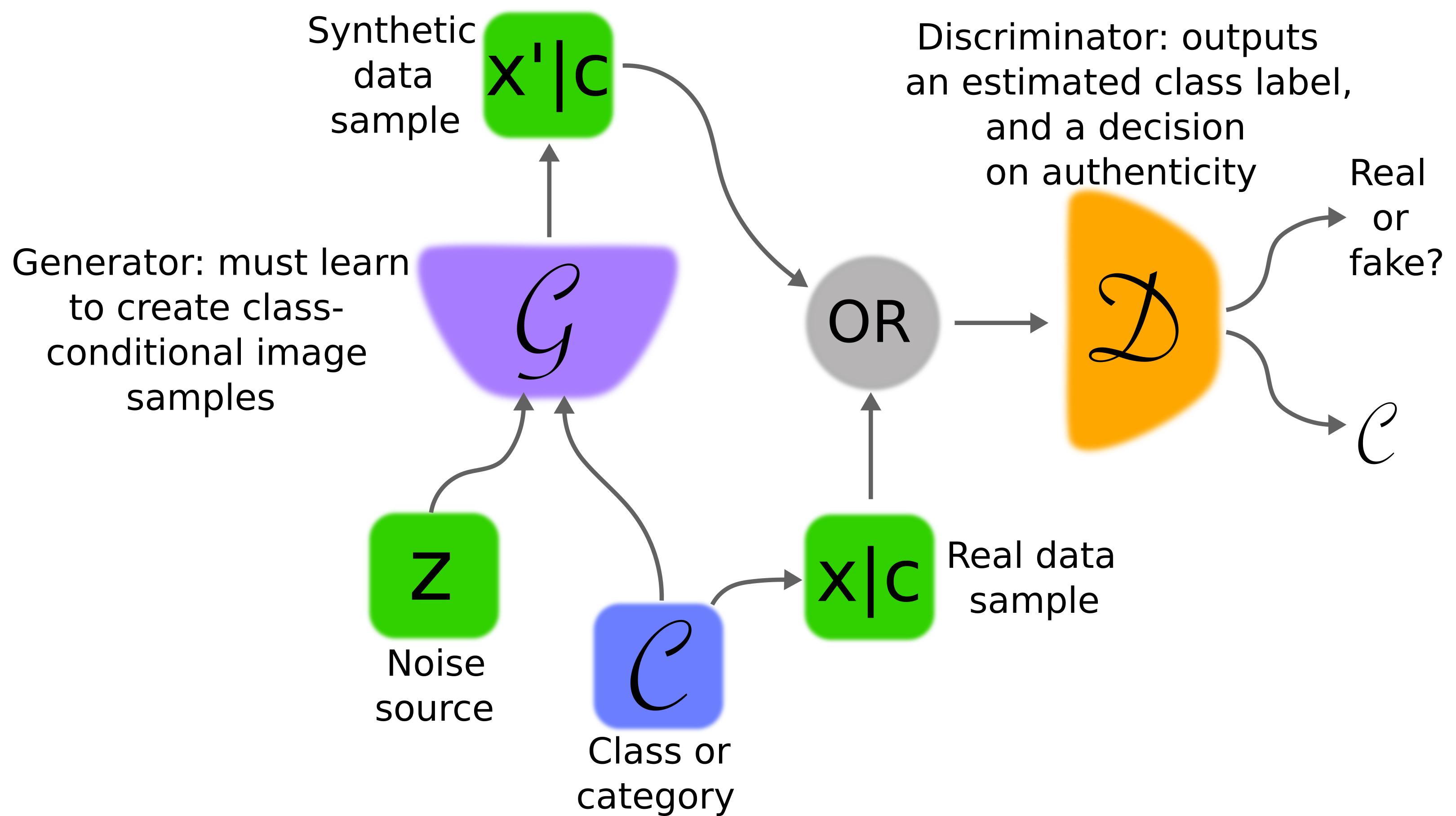}
\caption{Left, the Conditional GAN, proposed by Mirza et al. \cite{mirza2014conditional} performs class-conditional image synthesis; the discriminator performs class-conditional discrimination of real from fake images. The InfoGAN (right) \cite{chen2016infogan}, on the other hand, has a discriminator network that also estimates the class label.}
\label{fig:GANMiniZooPt1}
\end{figure*}

Mirza et al. \cite{mirza2014conditional} extended the (2D) GAN framework to the conditional setting by making both the generator and the discriminator networks class-conditional (Fig.~\ref{fig:GANMiniZooPt1}). Conditional GANs have the advantage of being able to provide better representations for multi-modal data generation. A parallel can be drawn between conditional GANs and InfoGAN \cite{chen2016infogan}, which decomposes the noise source into an incompressible source and a ``latent code'', attempting to discover latent factors of variation by maximizing the mutual information between the latent code and the generator's output. This latent code can be used to discover object classes in a purely unsupervised fashion, although it is not strictly necessary that the latent code be categorical. The representations learned by InfoGAN  appear to be semantically meaningful, dealing with complex inter-tangled factors in image appearance, including variations in pose, lighting and emotional content of facial images \cite{chen2016infogan}.

\subsection{GANs with Inference Models}
\label{sec:BiGANArch}
In their original formulation, GANs lacked a way to map a given observation, $\mathbf{x}$, to a vector in latent space -- in the GAN literature, this is often referred to as an inference mechanism. Several techniques have been proposed to invert the generator of pre-trained GANs \cite{creswell2016inverting,lipton2017precise}. The independently proposed Adversarially Learned Inference (ALI) \cite{dumoulin2016adversarially} and Bidirectional GANs \cite{donahue2016adversarial} provide simple but effective extensions,  introducing an inference network in which the discriminators examine joint $(\textrm{data},\textrm{latent})$ pairs. 

\begin{figure*}[h!]
\centering
\includegraphics[width=15cm]{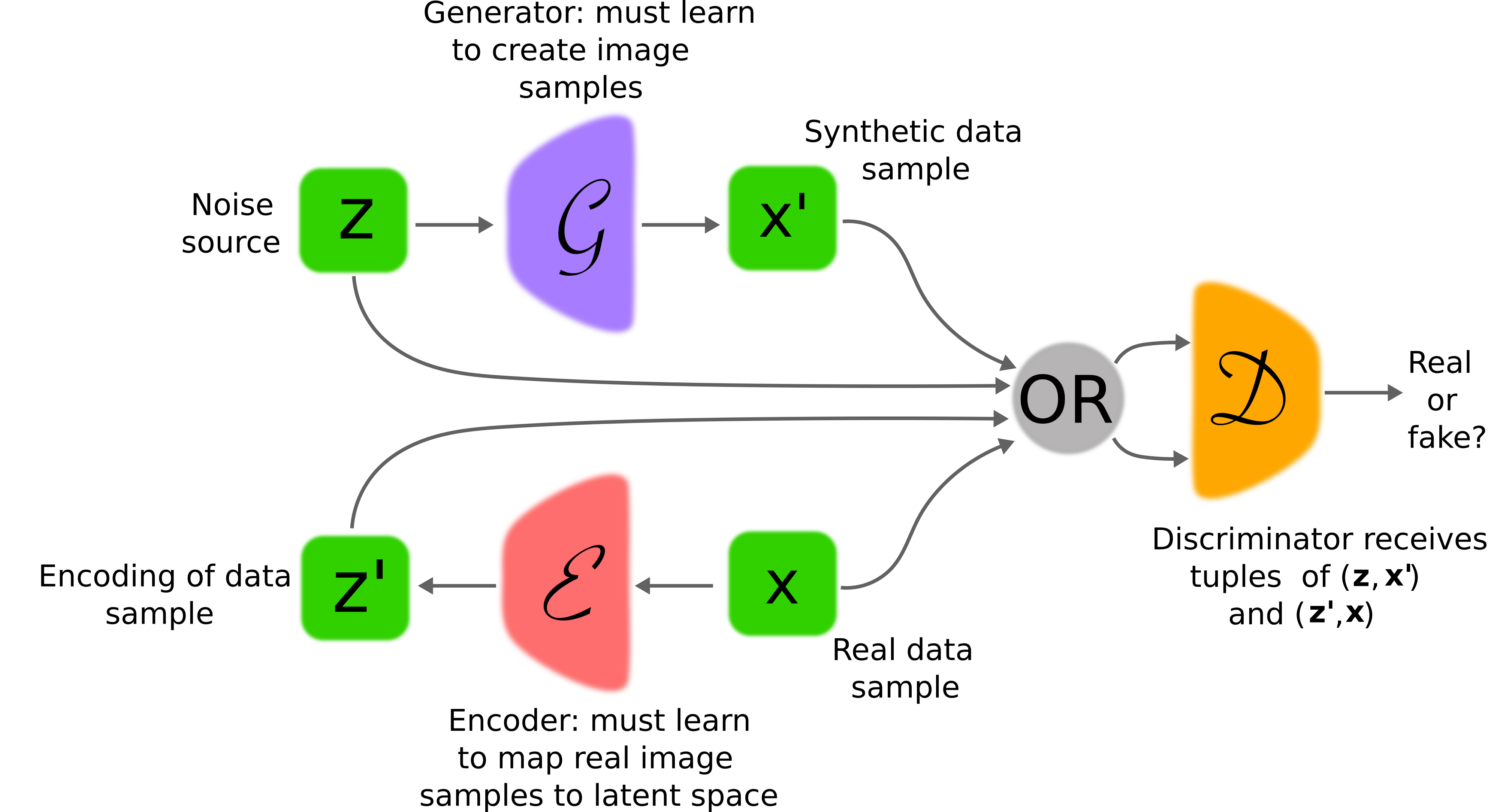}
\caption{The ALI/BiGAN structure \cite{donahue2016adversarial,dumoulin2016adversarially} consists of three networks. One of these serves as a discriminator, another maps the noise vectors from latent space to image space (decoder, depicted as a generator $\mathcal{G}$ in the figure), with the final network (encoder, depicted as $\mathcal{E}$) mapping from image space to latent space.}
\label{fig:BiGAN}
\end{figure*}

In this formulation, the generator consists of two networks: the ``encoder'' (inference network) and the ``decoder''. They are jointly trained to fool the discriminator. The discriminator itself receives pairs  of $(\mathbf{x}, \mathbf{z})$ vectors (see Fig.~\ref{fig:BiGAN}), and has to determine which pair constitutes a genuine tuple consisting of real image sample and its encoding, or a fake image sample and the corresponding latent-space input to the generator.


Ideally, in an encoding-decoding model the output, referred to as a reconstruction, should be similar to the input. Typically, the fidelity of reconstructed data samples synthesised using an ALI/BiGAN are poor. The fidelity of samples may be improved with an additional adversarial cost on the distribution of data samples and their reconstructions \cite{li2017towards}.

\subsection{Adversarial Autoencoders (AAE)}
\label{sec:AAEGANArch}
Autoencoders are networks, composed of an ``encoder'' and ``decoder'', that learn to map data to an internal latent representation and out again. That is, they learn a deterministic mapping (via the encoder) from a data space -- e.g., images -- into a latent or representation space, and a mapping (via the decoder) from the latent space back to data space.  The composition of these two mappings results in a ``reconstruction'', and the two mappings are trained such that a reconstructed image is as close as possible to the original. 

Autoencoders are reminiscent of the perfect-reconstruction filter banks that are widely used in image and signal processing. However, autoencoders generally learn non-linear mappings in both directions. Further, when implemented with deep networks, the possible architectures that can be used to implement autoencoders are remarkably flexible. Training can be unsupervised, with backpropagation being applied between the reconstructed image and the original in order to learn the parameters of both the encoder and the decoder.

As suggested earlier, one often wants the latent space to have a useful organization. Additionally, one may want to perform feedforward, ancestral sampling \cite{bengio2013generalized} from an autoencoder. Adversarial training provides a route to achieve these two goals. Specifically, adversarial training may be applied between the latent space and a desired prior distribution on the latent space (latent-space GAN). This results in a combined loss function \cite{makhzani2015adversarial} that reflects both the reconstruction error and a measure of how different the distribution of the prior is from that produced by a candidate encoding network. This approach is akin to a variational autoencoder (VAE) \cite{kingma2013auto} for which the latent-space GAN plays the role of the KL-divergence term of the loss function. 



Mescheder et al. \cite{Mescheder:2017} unified variational autoencoders with adversarial training in the form of the Adversarial Variational Bayes (AVB) framework. Similar ideas were presented in Ian Goodfellow's NIPS 2016 tutorial \cite{goodfellow2016nips}. AVB tries to optimise the same criterion as that of variational autoencoders, but uses an adversarial training objective rather than the Kullback-Leibler divergence. 

\section{Training GANs}
\label{sec:trainingGAN}
\subsection{Introduction}
\begin{figure*}[h!]
\centering
\includegraphics[width=12cm]{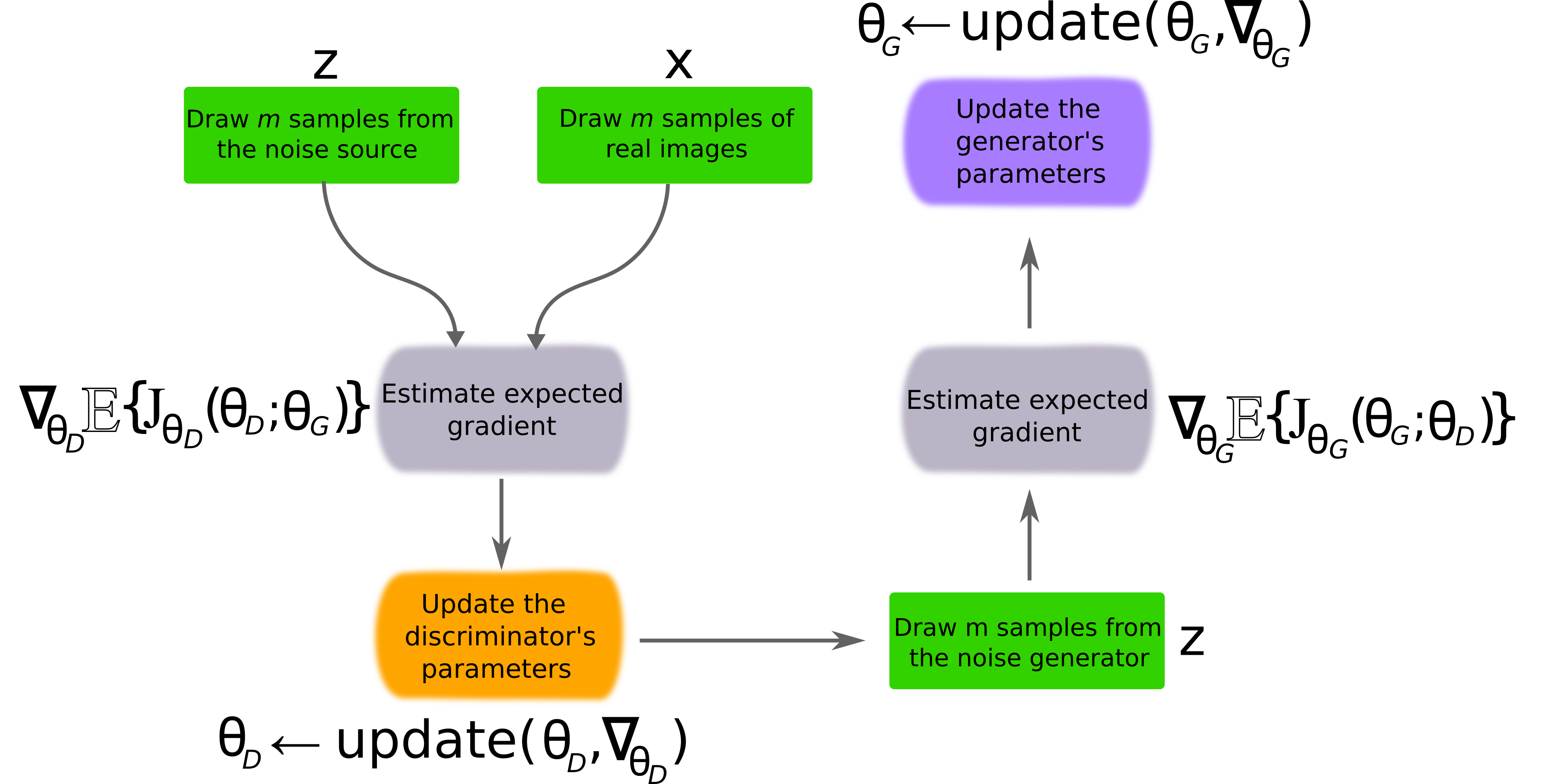}
\caption{The main loop of GAN training. Novel data samples, $\mathbf{x'}$, may be drawn by passing random samples, $\mathbf{z}$ through the generator network. The gradient of the discriminator may be updated $k$ times before updating the generator.} 
\label{fig:plaingancfg}
\end{figure*}

Training of GANs involves both finding the parameters of a discriminator that maximize its classification accuracy, and finding the parameters of a generator which maximally confuse the discriminator. This training process is summarized in Fig.~\ref{fig:plaingancfg}. 

The cost of training is evaluated using a value function, $V(\mathcal{G},\mathcal{D})$ that depends on both the generator and the discriminator. The training involves solving:

\[\max_\mathcal{D} \min_\mathcal{G} V(\mathcal{G},\mathcal{D}) \]
where
\[  V(\mathcal{G},\mathcal{D}) = \mathbb{E}_{p_{data}(\mathbf{x})} \log \mathcal{D}(\mathbf{x}) + \mathbb{E}_{p_g(\mathbf{x})} \log (1 - \mathcal{D}(\mathbf{x}))\]


During training, the parameters of one model are updated, while the parameters of the other are fixed. Goodfellow et al. \cite{goodfellow2014generative} show that for a fixed generator there is a unique optimal discriminator, $\mathcal{D}^*(\mathbf{x}) = \frac{p_{data}(\mathbf{x})}{p_{data}(\mathbf{x}) + p_g(\mathbf{x})}$. They also show that the generator, $\mathcal{G}$, is optimal when $p_g(\mathbf{x})=p_{data}(\mathbf{x})$, which is equivalent to the optimal discriminator predicting $0.5$ for all samples drawn from $\mathbf{x}$. In other words, the generator is optimal when the discriminator, $\mathcal{D}$, is maximally confused and cannot distinguish real samples from fake ones.

Ideally, the discriminator is trained until optimal with respect to the current generator; then, the generator is again updated. However in practice, the discriminator might not be trained until optimal, but rather may only be trained for a small number of iterations, and the generator is updated simultaneously with the discriminator. Further, an alternate, non-saturating training criterion is typically used for the generator, using $\max_\mathcal{G} \log \mathcal{D}(\mathcal{G}(\mathbf{z}))$ rather than $\min_\mathcal{G} \log (1 - \mathcal{D}(\mathcal{G}(\mathbf{z})))$.


Despite the theoretical existence of unique solutions, GAN training is challenging and often unstable for several reasons \cite{radford2015unsupervised}\cite{salimans2016improved}\cite{Arjovsky2016Towards}. One approach to improving GAN training is to asses the empirical ``symptoms'' that might be experienced during training. These symptoms include: 
\begin{itemize}
\item Difficulties in getting the pair of models to converge \cite{radford2015unsupervised};
\item The generative model, ``collapsing'', to generate very similar samples for different inputs \cite{salimans2016improved};
\item The discriminator loss converging quickly to zero \cite{Arjovsky2016Towards}, providing no reliable path for gradient updates to the generator.
\end{itemize}
Several authors suggested heuristic approaches to address these issues \cite{goodfellow2014generative,salimans2016improved}; these are discussed in Section \ref{sec:trainingTricks}.

Early attempts to explain why GAN training is unstable were proposed by Goodfellow and Salimans et al. \cite{goodfellow2014generative,salimans2016improved} who observed that gradient descent methods typically used for updating both the parameters of the generator and discriminator are inappropriate when the solution to the optimization problem posed by GAN training actually constitutes a saddle point. Salimans et al. provided a simple example which shows this \cite{salimans2016improved}. However, stochastic gradient descent is often used to update neural networks, and there are well developed machine learning programming environments that make it easy to construct and update networks using stochastic gradient descent. 

Although an early theoretical treatment \cite{goodfellow2014generative} showed that the generator is optimal when $p_g(\mathbf{x})=p_{data}(\mathbf{x})$, a very neat result with a strong underlying intuition, the real data samples reside on a manifold which sits in a high-dimensional space of possible representations. For instance, if colour image samples are of size $N\times N \times 3$ with pixel values $[0,\mathbb{R}^+]^3$, the space that may be represented -- which we can call $\mathbb{X}$ -- is of dimensionality $3N^2$, with each dimension taking values between 0 and the maximum measurable pixel intensity. The data samples in the support of $p_{data}$, however, constitute the manifold of the real data associated with some particular problem, typically occupying a very small part of the total space, $\mathbb{X}$. Similarly, the samples produced by the generator should also occupy only a small portion of $\mathbb{X}$.

Arjovsky et al. \cite{Arjovsky2016Towards} showed that the support $p_g(\mathbf{x})$ and $p_{data}(\mathbf{x})$ lie in a lower dimensional space than that corresponding to $\mathbb{X}$. The consequence of this is that $p_g(\mathbf{x})$ and $p_{data}(\mathbf{x})$ may have no overlap, and so there exists a nearly trivial discriminator that is capable of distinguishing  real samples, $\mathbf{x} \sim p_{data}(\mathbf{x})$ from fake samples, $\mathbf{x} \sim p_g(\mathbf{x})$ with $100\%$ accuracy. In this case, the discriminator error quickly converges to zero. Parameters of the generator may only be updated via the discriminator, so when this happens, the gradients used for updating parameters of the generator also converge to zero and so may no longer be useful for updates to the generator. Arjovsky et al.'s \cite{Arjovsky2016Towards} explanations account for several of the symptoms related to GAN training.

Goodfellow et al. \cite{goodfellow2014generative} also showed that when $\mathcal{D}$ is optimal, training $\mathcal{G}$ is equivalent to minimizing the Jensen-Shannon divergence between $p_g(\mathbf{x})$ and $p_{data}(\mathbf{x})$. If $\mathcal{D}$ is not optimal, the update may be less meaningful, or inaccurate. This theoretical insight has motivated research into cost functions based on alternative distances. Several of these are explored in Section \ref{alternativeCost}.

\subsection{Training Tricks} 
\label{sec:trainingTricks}

One of the first major improvements in the training of GANs for generating images were the DCGAN architectures proposed by Radford et al. \cite{radford2015unsupervised}. This work was the result of an extensive exploration of CNN architectures previously used in computer vision, and resulted in a set of guidelines for constructing and training both the generator and discriminator. In Section~\ref{sec:ConvGANArch}, we alluded to the importance of strided and fractionally-strided convolutions \cite{shelhamer2017fully}, which are key components of the architectural design. This allows both the generator and the discriminator to learn good up-sampling and down-sampling operations, which may contribute to improvements in the quality of image synthesis. More specifically to training, batch normalization \cite{ioffe2015batch} was recommended for use in both networks in order to stabilize training in deeper models. Another suggestion was to minimize the number of fully connected layers used to increase the feasibility of training deeper models. Finally, Radford et al. \cite{radford2015unsupervised} showed that using leaky ReLU activation functions between the intermediate layers of the discriminator gave superior performance over using regular ReLUs.


Later, Salimans et al. \cite{salimans2016improved} proposed further heuristic approaches for stabilizing the training of GANs. The first, feature matching, changes the objective of the generator slightly in order to increase the amount of information available. Specifically, the discriminator is still trained to distinguish between real and fake samples, but the generator is now trained to match the discriminator's expected intermediate activations (features) of its fake samples with the expected intermediate activations of the real samples. The second, mini-batch discrimination, adds an extra input to the discriminator, which is a feature that encodes the distance between a given sample in a mini-batch and the other samples. This is intended to prevent mode collapse, as the discriminator can easily tell if the generator is producing the same outputs. 


A third heuristic trick, heuristic averaging, penalizes the network parameters if they deviate from a running average of previous values, which can help convergence to an equilibrium. The fourth, virtual batch normalization, reduces the dependency of one sample on the other samples in the mini-batch by calculating the batch statistics for normalization with the sample placed within a reference mini-batch that is fixed at the beginning of training. 

Finally, one-sided label smoothing makes the target for the discriminator 0.9 instead of 1, smoothing the discriminator's classification boundary, hence preventing an overly confident discriminator that would provide weak gradients for the generator. S{\o}nderby et al. \cite{sonderby2017amortised} advanced the idea of challenging the discriminator by adding noise to the samples before feeding them into the discriminator. S{\o}nderby et al. \cite{sonderby2017amortised} argued that one-sided label smoothing biases the optimal discriminator, whilst their technique, instance noise, moves the manifolds of the real and fake samples closer together, at the same time preventing the discriminator easily finding a discrimination boundary that completely separates the real and fake samples. In practice, this can be implemented by adding Gaussian noise to both the synthesized and real images, annealing the standard deviation over time. The process of adding noise to data samples to stabilize training was, later, formally justified  by Arjovsky et al. \cite{Arjovsky2016Towards}.

\subsection{Alternative formulations} 
\label{alternativeCost}

The first part of this section considers other information-theoretic interpretations and generalizations of GANs. The second part looks at alternative cost functions which aim to directly address the problem of vanishing gradients.

\subsubsection{Generalisations of the GAN cost function}
Nowozin et al. \cite{nowozin2016f} showed that GAN training may be generalized to minimize not only the Jensen-Shannon divergence, but an \textbf{estimate} of $f$-divergences; these are referred to as $f$-GANs. The $f$-divergences include well-known divergence measures such as the Kullback-Leibler divergence. Nowozin et al. showed that $f$-divergence may be approximated by applying the Fenchel conjugates  of the desired $f$-divergence to samples drawn from the distribution of generated samples, after passing those samples through a discriminator \cite{nowozin2016f}. They provide a list of Fenchel conjugates for commonly used $f$-divergences, as well as activation functions that may be used in the final layer of the generator network, depending on the choice of $f$-divergence. Having derived the generalized cost functions for training the generator and discriminator of an $f$-GAN, Nowozin et al. \cite{nowozin2016f} observe that, in its raw form, maximizing the generator objective is likely to lead to weak gradients, especially at the start of training, and proposed an alternative cost function for updating the generator which is less likely to saturate at the beginning of training. This means that when the discriminator is trained, the derivative of the $f$-divergence on the ratio of the real and fake data distributions is estimated, while when the generator is trained only an estimate of the $f$-divergence is minimized. Uehara et al. \cite{uehara2016generative} extend the $f$-GAN further, where in the discriminator step the ratio of the distributions of real and fake data are predicted, and in the generator step the $f$-divergence is directly minimized. Alternatives to the JS-divergence are also covered by Goodfellow \cite{goodfellow2016nips}.

\subsubsection{Alternative Cost functions to prevent vanishing gradients}
Arjovsky et al. \cite{arjovsky2017wasserstein} proposed the WGAN, a GAN with an alternative cost function which is derived from an approximation of the Wasserstein distance. Unlike the original GAN cost function, the WGAN is more likely to provide gradients that are useful for updating the generator. The  cost function derived for the WGAN relies on the discriminator, which they refer to as the ``critic'', being a $k$-Lipschitz continuous function; practically, this may be implemented by simply clipping the parameters of the discriminator. However, more recent research \cite{gulrajani2017improved} suggested that weight clipping adversely reduces the capacity of the discriminator model, forcing it to learn simpler functions. Gulrajani et al. \cite{gulrajani2017improved} proposed an improved method for training the discriminator for a WGAN, by penalizing the norm of discriminator gradients with respect to data samples during training, rather than performing parameter clipping.

\subsection{A Brief Comparison of GAN Variants}
GANs allow us to synthesize novel data samples from random noise, but they are considered difficult to train due partially to vanishing gradients. All GAN models that we have discussed in this paper require careful hyperparameter tuning and model selection for training. However, perhaps the easier models to train are the AAE and the WGAN. The AAE is relatively easy to train because the adversarial loss is applied to a fairly simple distribution in lower dimensions (than the image data).  The WGAN \cite{gulrajani2017improved}, is designed to be easier to train, using a different formulation of the training objective which does not suffer from the vanishing gradient problem. The WGAN may also be trained successfully even without batch normalisation; it is also less sensitive to the choice of non-linearities used between convolutional layers.


Samples synthesised using a GAN or WGAN may belong to any class present in the training data. Conditional GANs provide an approach to synthesising samples with user specified content. 

It is evident from various visualisation techniques (Fig.~\ref{fig:smilevector}) that the organisation of the latent space harbours some meaning, but vanilla GANs do not provide an inference model to allow data samples to be mapped to latent representations. Both BiGANs and ALI provide a mechanism to map image data to a latent space (inference), however, reconstruction quality suggests that they do not necessarily faithfully encode and decode samples. A very recent development shows that ALI may recover encoded data samples faithfully \cite{li2017towards}. However, this model shares a lot in common with the AVB and AAE. These are autoencoders, similar to variational autoencoders (VAEs), where the latent space is regularised using adversarial training rather than a KL-divergence between encoded samples and a prior.

\section{The Structure of Latent Space}
\label{sec:SSoLS}
GANs build their own representations of the data they are trained on, and in doing so produce structured geometric vector spaces for different domains. This is a quality shared with other neural network models, including VAEs \cite{kingma2013auto}, as well as linguistic models such as \texttt{word2vec} \cite{mikolov2013efficient}. In general, the domain of the data to be modelled is mapped to a vector space which has fewer dimensions than the data space, forcing the model to discover interesting structure in the data and represent it efficiently. This latent space is at the ``originating'' end of the generator network, and the data at this level of representation (the latent space) can be highly structured, and may support high level semantic operations  \cite{radford2015unsupervised}. Examples include rotation of faces from trajectories through latent space, as well as image analogies which have the effect of adding visual attributes such as eyeglasses on to a ``bare'' face. 

\begin{figure*}[h!]
\centering
\includegraphics[width=12cm]{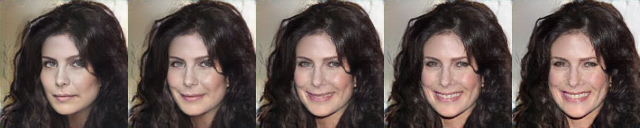}
\caption{Example of applying a ``smile vector'' with an ALI model \cite{dumoulin2016adversarially}. On the left hand side is an example of a woman without a smile and on the right a woman with a smile. A $\mathbf{z}$ value for the image of the woman on the left is inferred, $\mathbf{z}_1$ and for the right, $\mathbf{z}_2$. Interpolating along a vector that connects  $\mathbf{z}_1$ and $\mathbf{z}_2$, gives $\mathbf{z}$ values that may be passed through a generator to synthesize novel samples. Note the implication: a displacement vector in latent space traverses smile ``intensity'' in image space.}
\label{fig:smilevector}
\end{figure*}

All (vanilla) GAN models have a generator which maps data from the latent space into the space to be modelled, but many GAN models have an ``encoder'' which additionally supports the inverse mapping \cite{dumoulin2016adversarially, donahue2016adversarial}. This becomes a powerful method for exploring and using the structured latent space of the GAN network. With an encoder, collections of labelled images can be mapped into latent spaces and analysed to discover ``concept vectors'' that represent high level attributes such as ``smiling'' or ``wearing a hat''. These vectors can be applied at scaled offsets in latent space to influence the behaviour of the generator (Fig.~\ref{fig:smilevector}). Similar to using an encoding process to model the distribution of latent samples, Gurumurthy et al. \cite{gurumurthy2017deligan} propose modelling the latent space as a mixture of Gaussians and learning the mixture components that maximize the likelihood of generated data samples under the data generating distribution.


\section{Applications of GANs}
\label{sec:Apps}

Discovering new applications for adversarial training of deep networks is an active area of research. We examine a few computer vision applications that have appeared in the literature and have been subsequently refined. These applications were chosen to highlight some different approaches to using GAN-based representations for image-manipulation, analysis or characterization, and do not fully reflect the potential breadth of application of GANs.

Using GANs for image classification places them within the broader context of machine learning and provides a useful quantitative assessment of the features extracted in unsupervised learning. Image synthesis remains a core GAN capability, and is especially useful when the generated image can be subject to pre-existing constraints. Super-resolution \cite{ledig2016photo,yu2016ultra,yu2017hallucinating} offers an example of how an existing approach can be supplemented with an adversarial loss component to achieve higher quality results. Finally, image-to-image translation demonstrates how GANs offer a general purpose solution to a family of tasks which require automatically converting an input image into an output image.

\subsection{Classification and Regression}
After GAN training is complete, the neural network can be reused for other downstream tasks. For example, outputs of the convolutional layers of the discriminator can be used as a feature extractor, with simple linear models fitted on top of these features using a modest quantity of $(\textrm{image},\textrm{label})$ pairs \cite{radford2015unsupervised,salimans2016improved}. The quality of the unsupervised representations within a DCGAN network have been assessed by applying a regularized L2-SVM classifier to a feature vector extracted from the (trained) discriminator \cite{radford2015unsupervised}. Good classification scores were achieved using this approach on both supervised and semi-supervised datasets, even those that were disjoint from the original training data.


The quality of the data representation may be improved when adversarial training includes jointly learning an inference mechanism such as with an ALI \cite{dumoulin2016adversarially}. A representation vector was built using last three hidden layers of the ALI encoder, a similar L2-SVM classifier, yet achieved a misclassification rate significantly lower than the DCGAN \cite{dumoulin2016adversarially}. Additionally, ALI has achieved state-of-the art classification results when label information is incorporated into the training routine.

When labelled training data is in limited supply, adversarial training may also be used to synthesize more training samples. Shrivastava et al. \cite{shrivastava2016learning} use GANs to refine synthetic images, while maintaining their annotation information. By training models only on GAN-refined synthetic images (i.e. no real training data) Shrivastava et al. \cite{shrivastava2016learning} achieved state-of-the-art performance on pose and gaze estimation tasks. Similarly, good results were obtained for gaze estimation and prediction using a spatio-temporal GAN architecture \cite{zhang2017deep}. In some cases, models trained on synthetic data do not generalize well when applied to real data \cite{bousmalis2016unsupervised}. Bousmalis et al. \cite{bousmalis2016unsupervised} propose to address this problem by adapting synthetic samples from a source domain to match a target domain using adversarial training. Additionally, Liu et al. \cite{liu2016coupled} propose using multiple GANs -- one per domain -- with tied weights to synthesize pairs of corresponding images samples from different domains.

Because the quality of generated samples is hard to quantitatively judge across models, classification tasks are likely to remain an important quantitative tool for performance assessment of GANs, even as new and diverse applications in computer vision are explored.

\subsection{Image Synthesis}
Much of the recent GAN research focuses on improving the quality and utility of the image generation capabilities. The LAPGAN model introduced a cascade of convolutional networks within a Laplacian pyramid framework to generate images in a coarse-to-fine fashion \cite{denton2015deep}. A similar approach is used by Huang et al. \cite{huang2016stacked} with GANs operating on intermediate representations rather than lower resolution images. 

LAPGAN also extended the conditional version of the GAN model where both $\mathcal{G}$ and $\mathcal{D}$ networks receive additional label information as input; this technique has proved useful and is now a common practice to improve image quality. This idea of GAN conditioning was later extended to incorporate natural language. For example, Reed et al. \cite{reed2016generative} used a GAN architecture to synthesize images from text descriptions, which one might describe as {\it reverse captioning}. For example, given a text caption of a bird such as ``white with some black on its head and wings and a long orange beak'', the trained GAN can generate several plausible images that match the description. 

\begin{figure*}[h!]
\centering
\includegraphics[width=6cm]{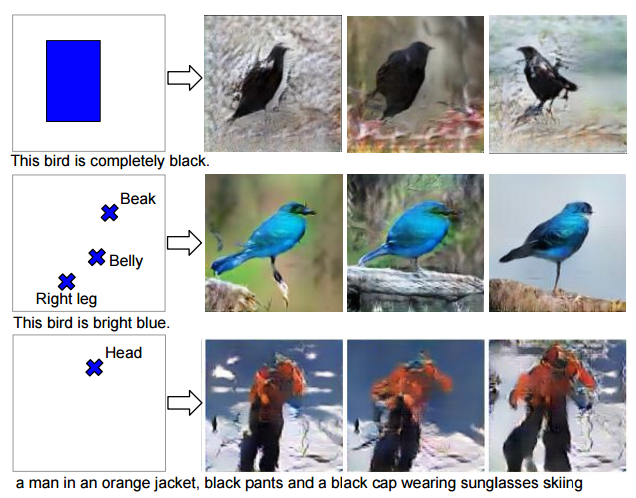}
\caption{Examples of Image Synthesis using the the Generative Adversarial What-Where Network (GAWWN). In GAWWN, images are conditioned on both text descriptions and image location specified as either by keypoint or bounding box. Figure reproduced from \cite{reed2016learning} with authors' permission.}
\label{fig:Sec6:A}
\end{figure*}

In addition to conditioning on text descriptions, the Generative Adversarial What-Where Network (GAWWN) conditions on image location \cite{reed2016learning}. The GAWWN system supported an interactive interface in which large images could be built up incrementally with textual descriptions of parts and user-supplied bounding boxes (Fig.~\ref{fig:Sec6:A}).

Conditional GANs not only allow us to synthesize novel samples with specific attributes, they also allow us to develop tools for intuitively editing images -- for example editing the hair style of a person in an image, making them wear glasses or making them look younger \cite{gurumurthy2017deligan}. Additional applications of GANs to image editing include work by Zhu and Brock et al. \cite{zhu2016generative, brock2016neural}.



\subsection{Image-to-image translation}
Conditional adversarial networks are well suited for translating an input image into an output image, which is a recurring theme in  computer graphics, image processing, and computer vision. The \texttt{pix2pix} model offers a general purpose solution to this family of problems \cite{isola2016image}. In addition to learning the mapping from input image to output image, the \texttt{pix2pix} model also constructs a loss function to train this mapping. This model has demonstrated effective results for different problems of computer vision which had previously required separate machinery, including semantic segmentation, generating maps from aerial photos, and colorization of black and white images. Wang et al. present a similar idea, using GANs to first synthesize surface-normal maps (similar to depth maps) and then map these images to natural scenes.

\begin{figure*}[h!]
\centering
\includegraphics[width=14cm]{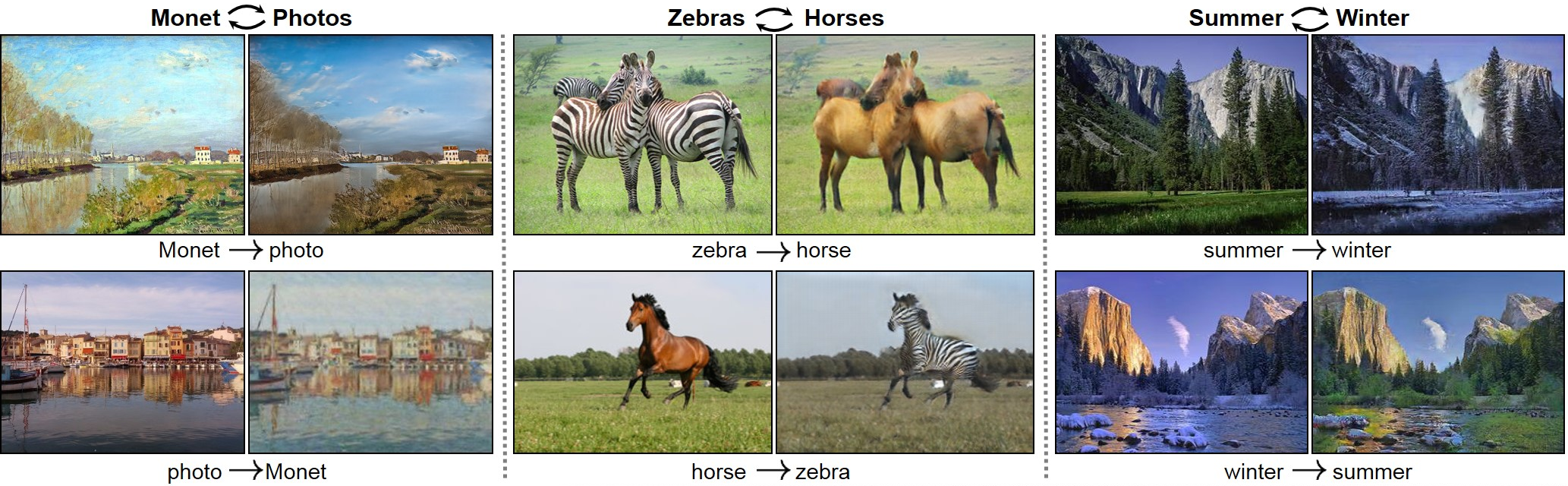}
\caption{CycleGAN model learns image to image translations between two unordered image collections. Shown here are the examples of bi-directional image mappings: Monet paintings to landscape photos, zebras to horses, and summer to winter photos in Yosemite park. Figure reproduced from \cite{zhu2017unpaired}.}
\label{fig:Sec6:B}
\end{figure*}

CycleGAN \cite{zhu2017unpaired} extends this work by introducing a cycle consistency loss that attempts to preserve the original image after a cycle of translation and reverse translation. In this formulation, matching pairs of images are no longer needed for training. This makes data preparation much simpler, and opens the technique to a larger family of applications. For example, artistic style transfer \cite{li2016precomputed} renders natural images in the style of artists, such as Picasso or Monet, by simply being trained on an unpaired collection of paintings and natural images (Fig.~\ref{fig:Sec6:B}). 

\subsection{Super-resolution} 
Super-resolution allows a high-resolution image to be generated from a lower resolution image, with the trained model inferring photo-realistic details while up-sampling. The SRGAN model \cite{ledig2016photo} extends earlier efforts by adding an adversarial loss component which constrains images to reside on the manifold of natural images.

The SRGAN generator is conditioned on a low resolution image, and infers photo-realistic natural images with 4x up-scaling factors. Unlike most GAN applications, the adversarial loss is one component of a larger loss function, which also includes perceptual loss from a pretrained classifier, and a regularization loss that encourages spatially coherent images. In this context, the adversarial loss constrains the overall solution to the manifold of natural images, producing perceptually more convincing solutions.

Customizing deep learning applications can often be hampered by the availability of relevant curated training datasets. However, SRGAN is straightforward to customize to specific domains, as new training image pairs can easily be constructed by down-sampling a corpus of high-resolution images. This is an important consideration in practice, since the inferred photo-realistic details that the GAN generates will vary depending on the domain of images used in the training set.

\section{Discussion}

\subsection{Open Questions}
GANs have attracted considerable attention due to their ability to leverage vast amounts of unlabelled data. While much progress has been made to alleviate some of the challenges related to training and evaluating GANs, there still remain several open challenges.

\subsubsection{Mode Collapse}
As articulated in Section~\ref{sec:trainingGAN}, a common problem of GANs involves the generator collapsing to produce a small family of similar samples (partial collapse), and in the worst case producing simply a single sample (complete collapse) \cite{Arjovsky2016Towards,arora2017generalization}.

Diversity in the generator can be increased by practical hacks to balance the distribution of samples produced by the discriminator for real and fake batches, or by employing multiple GANs to cover the different modes of the probability distribution \cite{Tolstikhin2017}. Yet another solution to alleviate mode collapse is to alter the distance measure used to compare statistical distributions. Arjovsky \cite{arjovsky2017wasserstein} proposed to compare distributions based on a Wasserstein distance rather than a KL-based divergence (DCGAN \cite{radford2015unsupervised}) or a total-variation distance (energy-based GAN \cite{zhao2016energy}). Metz et al. \cite{metz2016unrolled} proposed unrolling the discriminator for several steps, i.e., letting it calculate its updates on the current generator for several steps, and then using the ``unrolled'' discriminators to update the generator using the normal minimax objective. As normal, the discriminator only trains on its update from one step, but the generator now has access to how the discriminator would update itself. With the usual one step generator objective, the discriminator will simply assign a low probability to the generator's previous outputs, forcing the generator to move, resulting either in convergence, or an endless cycle of mode hopping. However, with the unrolled objective, the generator can prevent the discriminator from focusing on the previous update, and update its own generations with the foresight of how the discriminator would have responded.

\subsubsection{Training instability -- saddle points}

In a GAN, the Hessian of the loss function becomes indefinite. The optimal solution, therefore, lies in finding a saddle point rather than a local minimum. In deep learning, a large number of optimizers depend only on the first derivative of the loss function; converging to a saddle point for GANs requires good initialization. By invoking the stable manifold theorem from non-linear systems theory, Lee  et al. \cite{lee2016gradient} showed that, were we to select the initial points of an optimizer at random, gradient descent would not converge to a saddle with probability one (also see \cite{Pemantle1990,salimans2016improved}). Additionally, Mescheder et al. \cite{MeschederNG17a} have argued that convergence of a GAN's objective function suffers from the presence of a zero real part of the Jacobian matrix as well as eigenvalues with large imaginary parts.  This is disheartening for GAN training; yet, due to the existence of second-order optimizers, not all hope is lost. Unfortunately, Newton-type methods have compute-time complexity that scales cubically or quadratically with the dimension of the parameters. Therefore, another line of questions lies in applying and scaling second-order optimizers for adversarial training.

A more fundamental problem is the existence of an equilibrium for a GAN. Using results from Bayesian non-parametrics, Arora et al.  \cite{arora2017generalization} connects the existence of the equilibrium to a finite mixture of neural networks -- this means that below a certain capacity, no equilibrium might exist. On a closely related note, it has also been argued that whilst GAN training can appear to have converged, the trained distribution could still be far away from the target distribution. To alleviate this issue, Arora et al. \cite{arora2017generalization} propose a new measure called the `neural net distance'.

\subsubsection{Evaluating Generative Models}
How can one gauge the fidelity of samples synthesized by a generative models? Should we use a likelihood estimation? Can a GAN trained using one methodology be compared to another (model comparison)?  These are open-ended questions that are not only relevant for GANs, but also for probabilistic models, in general. Theis \cite{theis2016note} argued that evaluating GANs using different measures can lead conflicting conclusions about the quality of synthesised samples; the decision to select one measure over another depends on the application.

\subsection{Conclusions}
The explosion of interest in GANs is driven not only by their potential to learn deep, highly non-linear mappings from a latent space into a data space and back, but also by their potential to make use of the vast quantities of unlabelled image data that remain closed to deep representation learning. Within the subtleties of GAN training, there are many opportunities for developments in theory and algorithms, and with the power of deep networks, there are vast opportunities for new applications.

\section*{Acknowledgment}

The authors would like to thank David Warde-Farley for his valuable feedback on previous revisions of the paper. Antonia Creswell acknowledges the support of the EPSRC through a Doctoral training scholarship. 
\ifCLASSOPTIONcaptionsoff
  \newpage
\fi



\bibliographystyle{IEEEtran}
\bibliography{./bibtex/bib/thebib}
%
%

\begin{IEEEbiographynophoto}{Antonia Creswell} (ac2211@ic.ac.uk) holds a first-class degree from Imperial College in Biomedical Engineering (2011), and is currently a PhD student in the Biologically Inspired Computer Vision (BICV) Group at Imperial College London (2015). The focus of her PhD is on improving the training of generative adversarial networks and applying them to visual search and to learning representations in unlabelled sources of image data.
\end{IEEEbiographynophoto}

\begin{IEEEbiographynophoto}{Tom White}
Tom received his BS in Mathematics from the University of University of Georgia, USA, and MS from Massachusetts Institute of Technology in Media Arts and Sciences. He is currently a senior lecturer in the School of Design at Victoria University of Wellington, New Zealand. His current research focuses on exploring the growing use of constructive machine learning in computational design and the creative potential of human designers working collaboratively with artificial neural networks during the exploration of design ideas and prototyping.
\end{IEEEbiographynophoto}

\begin{IEEEbiographynophoto}{Vincent Dumoulin} holds a BSc in Physics and Computer Science from the University of Montr{\'e}al. He is a doctoral candidate at the Montr{\'e}al Institute for Learning Algorithms under the co-supervision of Yoshua Bengio and Aaron Courville, working on deep learning approaches to generative modelling.
\end{IEEEbiographynophoto}

\begin{IEEEbiographynophoto}{Kai Arulkumaran}
(ka709@ic.ac.uk) is a Ph.D. candidate in the
Department of Bioengineering at Imperial College London. He received a B.A. in Computer Science at the University of Cambridge in 2012, and an M.Sc. in Biomedical Engineering at Imperial College London in 2014. He was a Research Intern in Twitter Magic Pony and Microsoft Research in 2017. His research focus is deep reinforcement learning and computer vision for visuomotor control.
\end{IEEEbiographynophoto}

\begin{IEEEbiographynophoto}%
{Biswa Sengupta}
received his B.Eng. (Hons.) and M.Sc. degrees in electrical and computer engineering (2004) and  theoretical computer science (2005) respectively from the University of York.  He then read for a second M.Sc. degree in neural and behavioural sciences (2007) at the Max Planck Institute for Biological Cybernetics, obtaining his PhD in theoretical neuroscience (2011) from the University of Cambridge. He received further training in Bayesian statistics and differential geometry at the University College London and University of Cambridge before leading Cortexica Vision Systems as its Chief Scientist. Currently, he is a visiting scientist at Imperial College London along with leading machine learning research at Noah's Ark Lab of Huawei Technologies UK.  
\end{IEEEbiographynophoto}

\begin{IEEEbiographynophoto}{Anil A Bharath} (a.bharath@imperial.ac.uk) Anil Anthony Bharath is a Reader in the Department of Bioengineering at Imperial College London, an Academic Fellow of Imperial's Data Science Institute and a Fellow of the Institution of Engineering and Technology. He received a B.Eng. in Electronic and Electrical Engineering from University College London in 1988, and a Ph.D. in Signal Processing from Imperial College London in 1993. He was an academic visitor in the Signal Processing Group at the University of Cambridge in 2006. He is a co-founder of Cortexica Vision Systems. His research interests are in deep architectures for visual inference.

\end{IEEEbiographynophoto}




\end{document}